\documentclass[10pt, conference, final]{IEEEtran}
\usepackage{tikz}
\usepackage{pgfplots}
\usepackage{pgfplotstable}
\usepackage{tikz}
\usepackage{pgfplotstable}
\usepackage{amsmath,amssymb,amsfonts}
\usepackage{graphicx}
\usepackage{textcomp}
\usepackage{xcolor}
\usepackage{svg}
\usepackage{subcaption}
\renewcommand{\baselinestretch}{0.96} 

\usepackage[ruled,vlined,linesnumbered,noresetcount]{algorithm2e}
\SetKwFor{For}{for (}{) $\lbrace$}{$\rbrace$}

\usepackage{algpseudocode} 
\usepackage{siunitx,booktabs} %table
\usepackage{color,soul} 
\usepackage{xcolor}
\usepackage{flushend}
\usepackage{graphicx}
\usepackage{amssymb}
\usepackage{amsmath}
\usepackage{pdfpages}
\usepackage{lineno}
\usepackage{booktabs}
\usepackage{pifont}
\usepackage{tabu}
\usepackage{rotating}
\usepackage{array}
\usepackage[acronym,nomain]{glossaries}       
\usepackage{xcolor}
\usetikzlibrary{intersections,calc}
\usepackage{pgfplots}
\tikzset{
    axis break gap/.initial=1mm
}
\newglossary[slg]{symbolslist}{syi}{syg}{Symbolslist} % create add. symbolslist
\makeglossaries                                   
\usepackage{breqn}
\usepackage{booktabs}
\usepackage{soul}
\usepackage{tikz}
\usepackage{amsmath,amssymb,amsfonts}
\usepackage{graphicx}
\usepackage{colortbl}
\usepackage{multirow}
\usepackage[numbers,super]{}
\usepackage{pgf}
\usepackage{tikz,pgfplots}

\usepgfplotslibrary{groupplots,units}
\usetikzlibrary{patterns}
\usetikzlibrary{backgrounds,automata}
\pgfplotsset{compat=1.11,
    /pgfplots/ybar legend/.style={
    /pgfplots/legend image code/.code={%
       \draw[##1,/tikz/.cd,yshift=-0.25em]
        (0cm,0cm) rectangle (5pt,1.2em);},
   },
}

\newcommand{\blue}{\textcolor{blue}}

\definecolor{asparagus}{rgb}{0.53, 0.66, 0.42}
\definecolor{amber}{rgb}{1.0, 0.49, 0.0}
\definecolor{azure(colorwheel)}{rgb}{0.0, 0.5, 1.0}
\definecolor{azure2}{rgb}{0.0, 0.5, 1.0}

\DeclareMathOperator*{\argmin}{argmin}

\begin{document}
\title{Hybrid Learning for Orchestrating Deep Learning Inference in Multi-user Edge-cloud Networks}
\author{\IEEEauthorblockN{Sina Shahhosseini\textsuperscript{1}, Tianyi Hu\textsuperscript{1}, Dongjoo Seo\textsuperscript{1}, Anil Kanduri\textsuperscript{2}, Bryan Donyanavard\textsuperscript{3}, Amir M. Rahmani\textsuperscript{1}, Nikil Dutt\textsuperscript{1}}
\IEEEauthorblockA{\textit{\textsuperscript{1}  University of California, Irvine, \textsuperscript{2} University of Turku, Finland, \textsuperscript{3} San Diego State University} \\
 \{sshahhos, tianyih7, dseo3, a.rahmani, dutt\}@uci.edu, spakan@utu.fi,  bdonyanavard@sdsu.edu}}
\maketitle
\begin{abstract}
Deep-learning-based intelligent services have become prevalent in cyber-physical applications including smart cities and health-care. Collaborative end-edge-cloud computing for deep learning provides a range of performance and efficiency that can address application requirements through computation offloading. 
The decision to offload computation is a communication-computation co-optimization problem that varies with both system parameters (e.g., network condition) and workload characteristics (e.g., inputs). 
Identifying optimal orchestration considering the cross-layer opportunities and requirements in the face of varying system dynamics is a challenging multi-dimensional problem. While Reinforcement Learning (RL) approaches have been proposed earlier, they suffer from a large number of trial-and-errors during the learning process resulting
in excessive time and resource consumption. We present a Hybrid Learning orchestration framework that reduces the number of interactions with the system environment by combining model-based and model-free reinforcement learning. 
Our Deep Learning inference orchestration strategy employs reinforcement learning to find the optimal orchestration policy. Furthermore, we deploy Hybrid Learning (HL) to accelerate the RL learning process and reduce the number of direct samplings. We demonstrate efficacy of our HL strategy through experimental comparison with state-of-the-art RL-based inference orchestration, demonstrating that our HL strategy accelerates the learning process by up to 166.6$\times$. 
%We evaluate our HL strategy by comparing it with the state-of-the-art RL-based inference orchestration work. Our evaluation shows the HL strategy accelerates the learning process by up to $11.6\times$.
\end{abstract}
% \begin{IEEEkeywords}

% \end{IEEEkeywords}
\section{Introduction}

Deep-learning (DL) kernels provide intelligent end-user services in application domains such as computer vision, natural language processing, autonomous vehicles, and healthcare~\cite{schmidhuber2015deep}. 
End-user mobile devices are resource-constrained and rely on cloud infrastructure to handle the compute intensity of DL kernels \cite{wang2020convergence}. 
Unreliable network conditions and communication overhead in transmitting data from end-user devices affect real-time delivery of cloud services~\cite{khelifi2018bringing}. 
Edge computing brings compute capacity closer to  end-user devices, and complement the cloud infrastructure in providing low latency services \cite{yousefpour2018all}. 
Collaborative end-edge-cloud (EEC) architecture enables on-demand computational offloading of DL kernels from resource-constrained end-user devices to resourceful edge and cloud nodes \cite{shahhosseini2021exploring,shahhosseini2019dynamic}. 
Orchestrating DL services in multi-layered EEC architecture primarily focus on i) selecting an edge node onto which a task can be offloaded, and ii) selecting an appropriate learning model to accomplish the DL task. 
%rily requires selection of an edge and/or cloud node(s) onto which a specific task can be offloaded from the end-user devices, to minimize the overall latency. 
Selection of the edge node for offloading a DL task is based on a combination of factors including  i) the edge node's compute capacity and core-level heterogeneity, ii) communication penalty incurred in offloading, iii) workload intensity of the task and accuracy constraints, and iv) run-time variations in connectivity, signal strength, user mobility, and interaction. 
%The EEC architecture thus exposes a wide range of choices in task level partitioning and node selection for orchestrating DL tasks.
%The EEC architecture exposes different orchestration choices for i) selecting a node - based on node/core-level heterogeneity, and ii) selecting a task - based on compute, memory, and communication requirements. 
Selecting an appropriate model for a DL task depends on design time accuracy constraints and run-time latency constraints simultaneously. 
Different learning models for DL tasks expose a Pareto-space of accuracy-compute intensity, such that higher accuracy models consume longer execution time \cite{wang2020convergence}. 

Orchestrating DL tasks by finding the appropriate edge node for offloading, and configuring the learning model for DL task, while minimizing the latency under unpredictable network conditions makes orchestration a multi-dimensional optimization problem~\cite{kim2020autoscale}. Therefore, the orchestration problem requires an intelligent run-time management to search through a wide configuration Pareto-space. %Varying system dynamics at run-time adds to the complexity of orchestration strategies in being adaptable at run-time. Understanding run-time variations provide necessary feedback to make appropriate choices on system configurations such as offloading policies. 
%% Orchestrating DL tasks on multi-layered end-edge-cloud architectures , making it an NP-hard problem in the face of varying system dynamics. 
 Brute force search, heuristic, rule-based, and closed-loop feedback control solutions for orchestration require longer periods of time before converging to optimal decisions, making them inefficient for real-time orchestration \cite{sutton2018reinforcement}. Reinforcement Learning (RL) approaches have been adopted for orchestrating DL tasks in multi-layered end-edge-cloud systems~\cite{kim2020autoscale} to address these limitations. 
Orchestration strategies using RL can be classified into \textit{model-free} and \textit{model-based} approaches. 
\newline \textbf{Model-free RL} techniques operate with no assumptions about the system's dynamic or consequences of actions required to learn a policy. Model-free RL builds the policy model based on data collected through trial-and-error learning over epochs ~\cite{sutton2018reinforcement}. Existing \textit{Model-free} RL strategies have used Deep Reinforcement Learning (DRL) algorithms for minimizing the latency for multi-service nodes in end-edge-cloud architectures \cite{lu2020optimization}. AdaDeep~\cite{liu2020adadeep} proposes a resource-aware DL model selection using optimal learning. AutoScale~\cite{kim2020autoscale} proposes an energy-efficient computational offloading framework for DL inference. 
%Lu et. al. \cite{lu2020optimization} propose a Deep Reinforcement Learning (DRL) algorithm to minimize the latency for multi-service nodes in large-scale heterogeneous MEC and multi-dependence in mobile tasks. \cite{sen2019machine} proposes a Q-Learning framework to minimize energy by considering various parameters in task characteristics and resource availability. Young Geun et al. \cite{kim2020autoscale} propose a reinforcement learning based offloading technique for energy-efficient deep learning inference in the edge-cloud architecture. AdaDeep~\cite{liu2018demand} proposes a DRL algorithm to optimally select from a pool of compressed models according to available resources. 
Originated from trial-and-error learning, \textit{model-free} RL requires significant exploratory interactions with the environment \cite{sutton2018reinforcement}. Many of these interactions are impractical in distributed computer systems, since execution for each configuration is expensive and leads to higher latency and resource consumption during the learning process \cite{sutton2018reinforcement}. 
\newline \textbf{Model-based RL} uses a predictive internal model of the system to seek outcomes while avoiding the consequence of trial-and-error in real-time. Existing approaches have modeled the computation offloading problem and then use deep reinforcement learning to find an optimal orchestration solution \cite{zhan2020deep,lin2021computation} \cite{wang2019computation}. 
%Some efforts proposed MDP model-based solutions for the computation offloading problem and optimize it with DRL~\cite{zhan2020deep,lin2021computation}. Wang et al. \cite{wang2019computation} model the offloading problem and propose a model-based DRL to find an optimal offloading policy in different scenarios. 
{Model-based} RL approach is computationally efficient and provides better generalization and significantly less number of real full system execution runs before converging to the optimal solution \cite{sutton2018reinforcement}. However, {Model-based} RL is sensitive to model bias and suffers from model errors between the predicted and actual behavior leading to sub-optimal orchestration decisions.
\begin{table}[tb!]
\small
\centering
  \caption{State-of-the-art reinforcement learning-based orchestration frameworks for deep learning inference in end-edge-cloud networks. Approach- Model-free (MF) and Hybrid Learning (HL). Algorithm- Q-learning (QL), DeepQ (DQL), DeepDynaQ (DDQ). Actuation knobs- \textit{CO}: computation offloading, \textit{HW}: hardware knobs, \textit{APP}: application layer knobs. .}
  \resizebox{.45\textwidth}{!}{
  \begin{tabular}{ c c c c c}
    \toprule
     \parbox[c]{1.3cm}{\hrule height 0pt width 0pt \centering \textbf{Technique}} &
     \parbox[c]{1.2cm}{\hrule height 0pt width 0pt \centering \textbf{Approach}} &
     \parbox[c]{1.2cm}{\hrule height 0pt width 0pt \centering \textbf{Algorithm}}&
     \parbox[c]{1.2cm}{\hrule height 0pt width 0pt \centering \textbf{Workload}}&
     \parbox[c]{1cm}{\hrule height 0pt width 0pt \centering \textbf{Knobs}}\\
    \midrule
    \parbox[t]{1.3cm}{\centering AutoScale~\cite{kim2020autoscale}} & MF & QL & Inference & CO,HW\\ 
    \parbox[t]{1cm}{\centering AutoFL~\cite{kim2021autofl}}& MF & QL & Training & CO,HW\\
    \parbox[t]{1cm}{\centering AdaDeep~\cite{liu2020adadeep}}& MF &DQL&Inference& APP\\ 
    \parbox[t]{1cm}{\centering Ours}& HL &DDQ&Inference& CO,APP\\
    \bottomrule
  \end{tabular}
  }
  \label{table:related}
\end{table}

A hybrid learning approach integrating the advantages of both \textit{model-free} and \textit{model-based} RL is efficient for orchestrating DL tasks on end-edge-cloud architectures \cite{peng2018deep,sutton1990integrated}. 
In this work, we adapt such hybrid learning strategy for orchestrating deep learning tasks on distributed end-edge-cloud architectures. 
We model the end-edge-cloud system dynamics online, and design an RL agent that learns orchestration decisions. 
We incorporate the system model into the RL agent, which simulates the system and predicts the system behaviors. 
We exploit the hybrid \textit{Deep Dyna-Q}~\cite{peng2018deep,sutton1990integrated} model to design our RL agent, which requires fewer number of interactions with the end-to-end computer system, making the learning process efficient. 
Existing efforts to orchestrate DL inference/training over the network are summarized in Table~\ref{table:related}. 
We compare our framework with AutoScale~\cite{kim2020autoscale} and AdaDeep~\cite{liu2020adadeep} to demonstrate our agent's performance since they employ RL to optimally orchestrate DL inference at Edge.
Our main contributions are:
%It combines \textit{model-free} and \textit{model-based} RL by integrating planning, acting, and learning. 
%The strategy reduces the number of direct explorations and accelerates the learning process. In this paper, we will make the following
%With a hybrid strategy, the agent requires fewer number of interactions with the end-to-end computer system, making the learning process efficient.contributions:
%referred to as \textit{Deep Dyna-Q}~\cite{peng2018deep,sutton1990integrated},
\begin{itemize}
    \item A run-time reinforcement learning based orchestration framework for DL inference services, to minimize inference latency within accuracy constraints. 
    \item A hybrid learning approach to accelerate the RL learning process and  reduce direct sampling during  learning. 
    \item Experimental results demonstrating acceleration of the learning process on an end-edge-cloud platform by up to 166.6$\times$ over 
    state-of-the-art \textit{model-free} RL-based orchestration.
    %of our hybrid learning based orchestration strategy on an enterprise edge-cloud platform, by comparing against state-of-the-art \textit{model-free} RL-based orchestration.
\end{itemize}

\section{Online Learning for DL Inference Orchestration}
\label{OnlineLearningFramework}
%In this Section, we 
We begin by formulating the orchestration of DL inference on EEC architecture as an optimization problem, with the constraints of minimizing latency within an acceptable prediction accuracy. We design a RL agent to solve the orchestration problem, within the latency and accuracy constraints.

%Deep-learning (DL) is advancing real-time and interactive user services in domains such as autonomous vehicles, natural language processing, healthcare, and smart cities~\cite{schmidhuber2015deep}. 
%Due to user device resource constraints, deep learning kernels are often deployed on cloud infrastructure to meet computational demands~\cite{barbera2013offload}. 
%However, unpredictable network constraints including signal strength and delays affect real-time cloud services~\cite{khelifi2018bringing}. 
%Edge computing has emerged to complement cloud services, bringing compute capacity closer to the end-user devices \cite{yousefpour2018all}. The edge paradigm increases offloading opportunities for resource-constrained end-user devices.
%A collaborative end-edge-cloud architecture is essential to provide deep-learning-based services with acceptable latency to end-user devices \cite{mudassar2018edge}. Finding the optimal choice between offloading the DL tasks to edge and cloud layers and using optimized models for inference at local devices results in a high-dimensional optimization problem. In this work, we implement a DL inference orchestration framework which employs online reinforcement learning to orchestrate DL services for multi-users over the end-edge-cloud system. In the following we explain the framework:

\subsection{Problem Formulation}
%We define the optimization problem, then we employ a reinforcement learning agent to solve the problem. 
Consider an end-edge-cloud architecture, represented as (S,E,C), where %$S=\{S_1,S_2,..,S_5\}$ represents five end-device nodes; 
$S$, $E$, $C$ represents sensory device, edge, and cloud nodes respectively. Each edge and cloud node can service multiple sensory end device nodes. 
% \red{should we add a footnote to indicate the number 5 is just an example, but in principle parametrizable?}
In our model, we define $n$ as the number of sensory end-device such that $S=\{S_1,S_2,..,S_n\}$, representing $n$ end-device nodes. % we will explain that we evaluate our framework with up to five devices in the Experimental Setup section. 
Each of the $S$, $E$, $C$ nodes locally stores a pool of optimized inference models with different levels of computing intensity and model accuracy. 
Each end-device node runs an application that requires a DL inference task periodically. 
All end-device node resources are represented as a tuple $\{P_i,M_i,B_i\}$, where $P_i$ represents processor utilization of the node $i$; $M_i$ represents available memory at the node $i$; $B_i$ represents network connection status between the end-device node $i$ and edge and cloud nodes in higher layers. 
The computation offloading decision determines whether an end-device node should offload an inference task to a resourceful edge or cloud node, or perform the computation locally. 
The offloading decision for each end-device node is represented by $o_i$, for each end-device node $i$. 
The inference model selection determines the implementation of the model deployed for each inference on each end-node device. 
Each end-device node can perform inference with one of $l$ DL models $\{d_0,d_1,d_2,...,d_{l-1}\}$. 
In general, response time is the total time between making a request to a service and receiving the result. 
In our case, we define $T_{res}(o_i,d_k)$ as response time for a request from end-node device $i$ with offload decision $o_i$ and inference model $d_k$. 
Our objective is to minimize the average response time while satisfying the average accuracy constraint. 
% The problem is formulated in the following formula:
% \begin{equation}
% \begin{aligned}
% \textbf{P1:} \min_{}\quad & \frac{1}{N}\sum_{i=1}^{N} T_{res_i}(o_i,d_k)\\
%  %\textrm{s.t.}\quad & d_j \in \{d_1,d_2,d_3,...,d_l\}\\
%   \textrm{s.t.}\quad & \overline{accuracy}\;>\; threshold
% \end{aligned}
% \end{equation}
% where $\overline{accuracy}$ is the spatial average accuracy for simultaneous DL inferences.

% \subsection{Reinforcement Learning Agent}
\begin{table}[tb!]
\small
\caption{State Discrete Values}
\centering
\resizebox{\linewidth}{!}{
  \begin{tabular}{c c c}
    \toprule
     \parbox[c]{0.7cm}{\hrule height 0pt width 0pt \centering \textbf{State}}&
     \parbox[c]{2.1cm}{\hrule height 0pt width 0pt \centering \textbf{Discrete Values}}&
     \parbox[c]{2.1cm}{\hrule height 0pt width 0pt \centering \textbf{Description}}\\
    \midrule
    \parbox[c]{2.1cm}{\centering $P^{S_i}$}& Available, Busy & End-node CPU Utilization\\
    \parbox[c]{2.1cm}{\centering $M^{S_i}$}& Available, Busy& End-node Memory Utilization\\ 
    \parbox[c]{2.1cm}{\centering $B^{S_i}$}& Regular, Weak & End-node Available Bandwidth\\ 
    \parbox[c]{2.1cm}{\centering $P^{E}$}& Nine discrete levels & Edge CPU Utilization\\
    \parbox[c]{2.1cm}{\centering $M^{E}$}& Available, Busy & Edge Memory Utilization\\
    \parbox[c]{2.1cm}{\centering $B^{E}$}& Regular, Weak & Edge Available Bandwidth\\ 
    \parbox[c]{2.1cm}{\centering $P^{C}$}& Nine discrete levels & Cloud CPU Utilization\\
    \parbox[c]{2.1cm}{\centering $M^{C}$}& Available, Busy& Cloud Memory Utilization\\ 
    \parbox[c]{2.1cm}{\centering $B^{C}$}& Regular, Weak & Cloud Available Bandwidth\\ 
    \bottomrule
  \end{tabular}
  }
  \label{table:SD}
  \vspace{-2mm}
\end{table}

\subsection{RL Agent}
Reinforcement learning (RL) is widely used to automate intelligent decision making based on experience. Information collected over time is processed to formulate a policy which is based on a set of rules. 
Each rule consists of three major components: \textit{\textbf{(a) State Space:}} State describes the current system dynamics in terms of available cores, memory, and network resources. 
These entities affect the inference performance, hence, our state vector is composed of CPU utilization, available memory, and bandwidth per each computing resource. 
Table \ref{table:SD} shows the discrete values for each component of the state. 
\textit{\textbf{(b) Action Space:}} RL actions represent the orchestration knobs in the system. 
We define actions as the choice of compute node among available execution options at (S,E,C) layers, and choice of inference model to deploy. 
We limit the edge and cloud devices to always use the high accuracy inference model, and the end-node devices have a choice of $l$ different models. 
Therefore, the action space is defined as $a_{\tau}=\{o^i,d_j\}$ where $i \in \{S,E,C\}$ and $d_j \in \{d_1,d_2,...,d_l\}$.
% \begin{equation}
% \begin{split}
% S_{\tau}=\{P^{E},M^{E},B^{E},P^{C},M^{C},B^{C},P^{S_1},M^{S_1},B^{S_1},...,
% P^{S_n}\\,M^{S_n},B^{S_n}\}
% \end{split}
% \end{equation}
 \textit{\textbf{(c) Reward Function:}} A reward in RL is a feedback from the environment to optimize objective of the system. 
 In our work, the reward function is defined as the average response time of DL inference requests. 
 In our case, the agent seeks to minimize the average response time. 
 To ensure the agent minimizes the average response time while satisfying the accuracy constraint, we penalize the agent when the accuracy threshold is violated. 
 On the other hand, when the selected action satisfies the constraint, the reward is the average response time.

 \begin{figure}[tb!]
    \centering
    \includegraphics[width=0.4\textwidth]{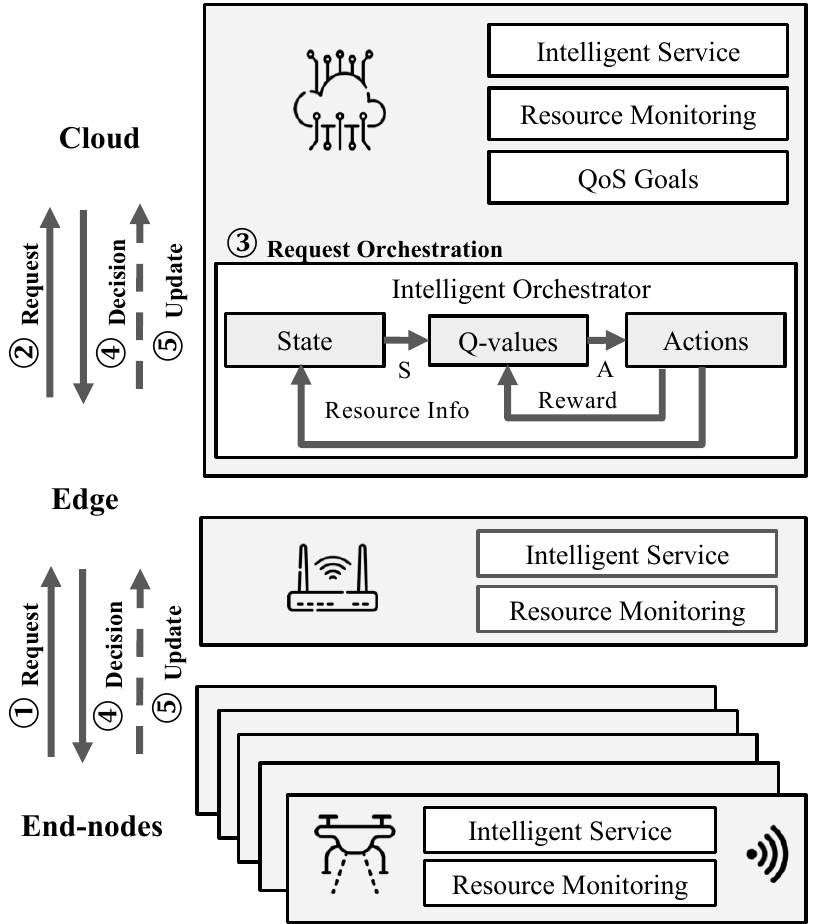}
    \caption{Online learning framework for orchestrating DL inference.}
    \label{fig:framework}
    \vspace{-6mm}
\end{figure}
\subsection{DL Inference Orchestration Framework}
Figure \ref{fig:framework} shows our end-edge-cloud architecture framework, integrating service requests, resource monitoring, and intelligent orchestration. 
The \textit{Intelligent Orchestrator} (IO) acts as an RL-agent for making computation offloading and model selection decisions. 
The end-device layer consists of multiple end-user devices. 
Each end-device has two software components: (i) \textit{Intelligent Service} - an image classification kernel with DL models of varying compute intensity and prediction accuracy; and
(ii) \textit{Resource Monitoring} - a periodic service that collects the device's system parameters including CPU and memory utilization, and network condition, and broadcasts the information to the edge and cloud layers.
Both the edge and cloud layers also have the \textit{Intelligent Service} and \textit{Resource Monitoring} components. 
The \textit{Intelligent Orchestrator} acts a centralized RL-agent that is hosted at the cloud layer for inference orchestration. 
The agent collects resource information including processor utilization, available memory, and network condition) from Resource Monitoring components throughout the network. 
The agent also gathers the reward value (i.e., response time) from the environment in order to learn an optimal policy. 
The \textit{Quality of Service Goal} provides the required QoS for the system (i.e., the accuracy constraint). Figure~\ref{fig:framework} illustrates a step-wise procedure of the inference service in our framework. 
The end-device layer consists of resource-constrained devices that periodically make requests for DL inference service (Step 1). 
The requests are passed through the edge layer (Step 2) to the cloud device to be processed by \textit{Intelligent Orchestrator} (Step 3). 
The agent determines where the computation should be executed, and delivers the \textit{Decision} to the network (Step 4). Each device updates the agent after it performs an inference with the response time information of the requested service (Step 5). 
In addition, all devices send the available resource information including the processor utilization, available memory, and network condition to the cloud device (Step 5).

% singh1992reinforcement
% sutton1996model

\section{Hybrid Learning Strategy}
\begin{figure}[t]
    \centering
    \includegraphics[width=0.48\textwidth]{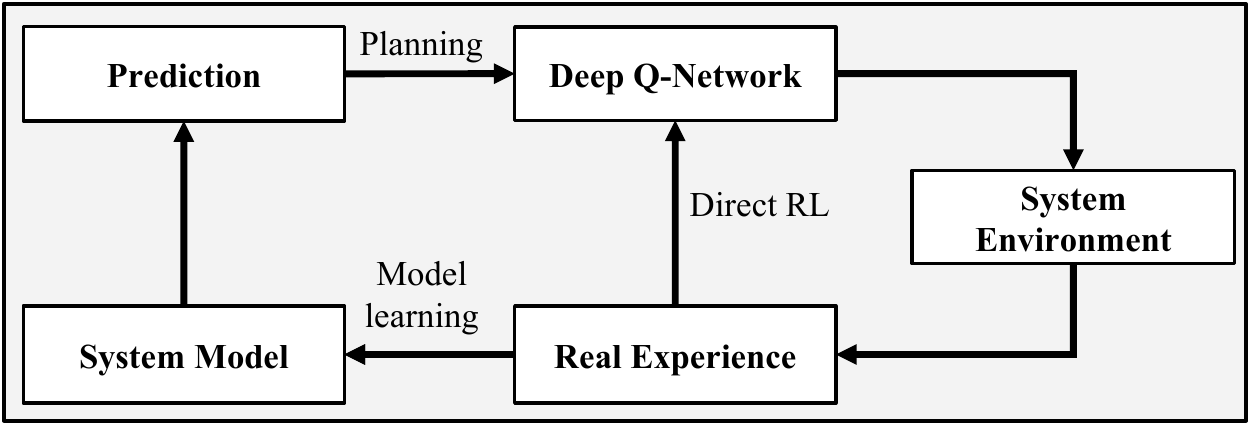}
    \caption{Hybrid Learning Architecture.}
    \vspace{-2mm}
    \label{fig:archhybridlearning}
\end{figure}

Hybrid Learning is a combination of \textit{model-free} and \textit{model-based} RL. 
As illustrated in Figure \ref{fig:archhybridlearning}, the architecture consists of \textit{System Environment}, \textit{System Model}, and \textit{Policy Model}. 
The training process begins with an initial system model and an initial policy. 
The agent is trained in three phases viz., \textbf{\textit{Direct RL}}, \textbf{\textit{System Model Learning}}, \textbf{\textit{Planning}}. 
Algorithm 1 defines the training process, which is composed of three major phases:
% : the agent interacts with System Environment to collect Real Experience to train Deep Q-Network model
% : the System Model consists of two models, which are supposed to \textit{Predict} the system's behavior for given pairs of current state and chosen action. These models are learned and updated through real experiences. 
% : the agent uses the System Model to predict system's behavior which helps to improve policy model.

\noindent\textbf{(1) Direct RL:} In the Direct-RL phase, the agent interacts with the \textit{System Environment} to collect Real Experience for training the Deep Q-Network (DQN) model. 
Every time the agent takes a \textit{step}, the \textit{real experience} is pushed into a prioritized replay buffer $D_{direct}$, and a random replay buffer $D_{world}$. 
We sample mini-batches from the buffer $D_{direct}$ and update the DQN by Adam optimizer~\cite{zhang2018improved}. 
Then, we assign new priorities to the prioritized replay buffer $D_{direct}$.
% \begin{equation}
% \label{equation:TD-Loss}
% \begin{aligned}
% \mathcal{L}\left(\theta_{Q}\right) &=\mathbb{E}_{\left(s_{\tau}, a_{\tau}, r, s_{\tau+1}\right) \in MiniBatch}\left[\left(y-Q\left(s_{\tau}, a_{\tau} ; \theta_{Q}\right)\right)^{2}\right] \\
% y &=r+\gamma \max _{a_{\tau+1}} Q^{\prime}\left(s_{\tau+1}, a_{\tau+1} ; \theta_{Q^{\prime}}\right)
% \end{aligned}
% \end{equation}

\noindent\textbf{(2) System Model Learning:} 
We model our system to \textit{Predict} the system's behavior for given pairs of ($s_\tau$,$a_\tau$). \textit{System Model Learning} starts with no assumption about the \textit{System Environment}, and is learned and updated through real experiences. As the agent takes more steps with real experiences, the model ($System(\theta_{s})$) learns the system from state-action pairs that have previously been experienced (in Direct RL). $System(s_\tau,a_\tau;\theta_{s})$ predicts average response time $r_{\tau}$ and the next state $s_{\tau+1}$. In this phase, we train the model with mini-batches sampled randomly from the buffer $D_{world}$ update the $\theta_s$ accordingly.

\noindent\textbf{(3) Planning:} 
% \iffalse
% the System Model is employed to generate predictions for random states to improve the policy model. If the model predicts accurately the system behavior, a number of iterative planning steps can significantly speed up the policy learning.
% \fi
During this phase, the agent uses the \textit{System Model} to predict the system's behavior to improve the policy model $Q(s,a;\theta_Q)$. We train $Q(s,a;\theta_Q)$ with the predicted tuples $(s_\tau,a_\tau,r_\tau,s_{\tau+1})$ in a replay buffer $D_{plan}$. 
Given a current state $s$, we use $System(s,a; \theta_s)$ to generate a set $A$ of $K$ actions ($a_i, i = 0,\cdots,K-1$) that might yield promising rewards $r$. For each action $a_i$ in the set $A$, if $a_i$ does not exist in the buffer $D_{plan}$, agent will take a step $a_i$ at state $s$ to get a reward $r$ and the next state $s^{\prime}$. Then we push the tuple $(s, a_i, r, s^{\prime})$ into the buffer $D_{plan}$.
If the action $a_i$ already exists in buffer $D_{plan}$, we only update its corresponding current state in the buffer with the new state. 
% We utilize the fact that the reward $r$ is only related to the action $a$ instead of state $s$ in our reward system design so that we can avoid the expensive overhead of system environment interactions.
Then, we train the policy model $Q(s,a;\theta_{Q})$ in the same way as the Direct RL process but with the generated data sampled from the buffer $D_{plan}$.

We define $\alpha$ as a parameter to control the portion of \textit{Direct RL} and \textit{Planning} during the training policy. Increasing $\alpha$ over time results in decreasing the number of Direct RL during the training. In this strategy, after sufficient real experience, the \textit{System Model} can predict the system behavior. Therefore, the agent relies more on \textit{System Model} prediction rather than real experiences which emphasizes   \textit{model-based} RL.

\begin{algorithm}[h]
\small
% \algsetup{linenosize=\tiny}
\caption{Hybrid Learning Algorithm.}
\SetInd{0.3em}{0.3em}
\KwData{$\epsilon, C, T, N$}
\KwResult{$Q(s,a;\theta_{Q}), System(s,a;\theta_{s}))$}
initialize $Q(s,a;\theta_{Q})$ and $Q^{\prime}(s,a;\theta_{Q^{\prime}})$ with $\theta_{Q^{\prime}} \leftarrow \theta_{Q}$\;
% initialize $Time(s,a;\theta_{T})$ and $State(s,a;\theta_{S})$\;
initialize $System(s,a;\theta_{s})$\;
initialize replay buffer $D_{direct},\!D_{world}, \!D_{plan}$\;
\For{$epoch \leftarrow 1:N$}{
    $\alpha \leftarrow \frac{epoch}{N}$\;
    \blue{$\#$ Direct Reinforcement Learning ---------------------\;}
    \For{$session \leftarrow 1:(1-\frac{\alpha}{2})N_{direct}$}{
        % $s\leftarrow env.reset()$\;
        \For {$step \leftarrow 1:T_{direct}$}{
            %\blue{$\#$ Epsilon-Greedy Algorithm}
            with probability $\epsilon$, choose random $a$,
            otherwise $a \leftarrow \argmin_{a^{\prime}}Q(s,a^{\prime};\theta_Q)$\;
            $r, s^{\prime} \leftarrow take\;step (s,a)$\;
            store $(s,a,r,s^{\prime}) \rightarrow D_{direct}, D_{world}$\;
            $s \leftarrow s^{\prime}$\;
            sample prioritized minibatch $(S,A,R,S^{\prime})\!\subset\! D_{direct}$\;
            % update $\theta_{Q}$ via Equation \ref{equation:TD-Loss} on the minibatch\;
            update $\theta_{Q}$ via Adam on the minibatch\;

        }
    }
    update target model $Q^{\prime}(s,a;\theta_{Q^{\prime}})$ by $\theta_{Q^{\prime}} \leftarrow \theta_{Q}$\;
    
    \blue{$\#$ System Model ----------------------------------------\;}
    \For{$session \leftarrow 1:(1-\frac{\alpha}{2})N_{world}$}{
        sample random minibatch $(S,A,R,S^{\prime})\subset D_{world}$\;
        update $\theta_S$\;
        % update $\theta_T$ via Adam of the MSE loss\;
        % update $\theta_S$ via Adam of the cross-entropy loss\;
    }
    
    \blue{$\#$ Planning -------------------------------------------------\;}
    \For{$session \leftarrow 1:(\frac{\alpha+1}{2})N_{suggest}$}{
        % $s \leftarrow env.reset()$\;
        \For {$step \leftarrow 1:T_{suggest}$}{
            $a\!\leftarrow\! \mathop{\arg\!\min}_{a^{\prime}} System(s,a^{\prime};\theta_s)$\; 
            % $a\!\leftarrow\! \mathop{\arg\!\min}_{a^{\prime}} Time(s,a^{\prime};\theta_T)$\; 
            % $s^{\prime}\! \leftarrow\! State(s, a;\theta_S)$\;
            % $a\!\leftarrow\! \mathop{\arg\!\min}_{a^{\prime}} Time(s,a^{\prime};\theta_T)$\; 
            $s^{\prime}\! \leftarrow\! System(s, a;\theta_S)$\;
            \blue{$\#$ Choose best $K$ actions for current state $s$\;}
            $A\!\leftarrow\!\{a | a\!\in\! argsort_{a^{\prime}}(System(s,a^{\prime};\theta_s))[0\!:\!K]\}$\;
            \For{$a^{\prime}$ in $A$}{
                \eIf{$a^{\prime} \not\in D_{plan}$}{                        $r, s^{\prime} \leftarrow take\;step (s,a^{\prime})$\;
                    store $(s, a^{\prime}, r, s^{\prime}) \rightarrow D_{plan}$\;
                }{
                    update $\!(s^{\prime\prime},a^{\prime},r,s^{\prime})\!\in\! D_{plan}$ by $s^{\prime\prime} \leftarrow s$\;
                }
            }
            $s \leftarrow s^{\prime}$\;
        }
    }
    \For{$session \leftarrow 1:(\frac{\alpha+1}{2})N_{plan}$}{
        sample prioritized minibatch from $D_{plan}$\;
        update $\theta_Q$ via Adam on the minibatch\;
    %   update $\theta_Q$ via Adam of Equation \ref{equation:TD-Loss} on the minibatch\;

    }
    update target model $Q^{\prime}(s,a;\theta_{Q^{\prime}})$ by $\theta_{Q\prime} \leftarrow \theta_{Q}$\;
}
\end{algorithm}
\vspace{-4mm}
% \vspace{-25mm}
\section{Evaluation}
\subsection{Experimental Setup}
In this subsection, we describe the experimental setup for our proposed framework. First, we explain the DL workloads as benchmarks in our evaluations. Then, we explain the scenarios to evaluate our framework. Finally, we describe the platform setup.
For DL workloads, we consider the MobileNet image classification application as the benchmark  \cite{howard2017mobilenets}. We consider eight different MobileNet models~\cite{howard2017mobilenets} ($d0$ through $d7$) with varying levels of accuracy and performance. Table \ref{table:accuracy} summarizes the MobileNet models $d0$ through $d7$, with each model having different response time and accuracy levels. Our framework supports up to five end-device nodes, networked with edge and cloud layers. Each end-user device is connected to a single edge device, and can request a DL inference service to the cloud layer. The cloud layer hosts the IO that contains the RL agent, which handles the inference orchestration. Upon on each service request, the RL agent is invoked to determine: (i) where the request should be processed and (ii) what DL model should be executed for the corresponding request. The platform consists of five AWS a1.medium instances with single ARM-core (as the end-device nodes), connected to an AWS a1.large instance with two CPUs (as edge device), and an AWS a1.xlarge instance with four CPUs (as cloud node). We perform the training process using NVIDIA RTX 5000 at the cloud node. In this work, we conduct experiments under four unique scenarios with varying network conditions (See Table \ref{table:EE}). Each scenario represents a combination of regular (R) and weak (W) network signal strength over five end-user devices (S1-S5) and 1 edge device (E). The regular network has no transmission delay, while we add $20$ms delay to all outgoing packets for the weak connection.
  \begin{table}[tb!]
\small
\caption{MobileNet Models \cite{howard2017mobilenets}}
\centering
\resizebox{0.5\textwidth}{!}{
  \begin{tabular}{c c c  c c}
    \toprule
    \parbox[c]{0.5cm}{\hrule height 0pt width 0pt \centering \textbf{\#}}&
     \parbox[c]{2.7cm}{\hrule height 0pt width 0pt \centering \textbf{Model}}&
     \parbox[c]{2cm}{\hrule height 0pt width 0pt \centering \textbf{Million MACs}}&
     \parbox[c]{0.7cm}{\hrule height 0pt width 0pt \centering \textbf{Type}}&
     \parbox[c]{2cm}{\hrule height 0pt width 0pt \centering \textbf{Accuracy (\%)}}\\
    \midrule
    $d0$ &\parbox[c]{3cm}{\centering 1.0 MobileNetV1-224}& 569& FP32 &  89.9\\   
    $d1$&\parbox[t]{3cm}{\centering 0.75 MobileNetV1-224} & 317 & FP32 & 88.2\\
    $d2$&\parbox[t]{3cm}{\centering 0.5 MobileNetV1-224}& 150& FP32&84.9\\ 
    $d3$&\parbox[t]{3cm}{\centering 0.25 MobileNetV1-224} & 41& FP32&74.2\\ 
    $d4$&\parbox[t]{3cm}{\centering 1.0 MobileNetV1-224} & 569& Int8&88.9\\ 
    $d5$&\parbox[t]{3cm}{\centering 0.75 MobileNetV1-224}& 317& Int8&87.0\\ 
    $d6$&\parbox[t]{3cm}{\centering 0.5 MobileNetV1-224}&150& Int8& 83.2\\
    $d7$&\parbox[t]{3cm}{\centering 0.25 MobileNetV1-224} &41& Int8&72.8\\ 
    \bottomrule
  \end{tabular}
  }
\label{table:accuracy}
\end{table}

% \begin{table}[t!]
% \small
% \caption{Device Specification}
% \centering
% \resizebox{.4\textwidth}{!}{
%   \begin{tabular}{c c  c c}
%     \toprule
%      \parbox[c]{1.5cm}{\hrule height 0pt width 0pt \centering \textbf{Node Type}}&
%      \parbox[c]{0.7cm}{\hrule height 0pt width 0pt \centering \textbf{vCPUs}}&
%      \parbox[c]{1.5cm}{\hrule height 0pt width 0pt \centering \textbf{Memory (GiB)}}&
%      \parbox[c]{1.5cm}{\hrule height 0pt width 0pt \centering \textbf{Frequency (GHz)}}\\
%     \midrule
%     End & 1 & 2 & 2.3  \\   
%     Edge & 2 & 4 & 2.3 \\   
%     Cloud & 4 & 8 & 2.3 \\   
%     \bottomrule
%   \end{tabular}
%   }
%   \label{table:DS}
%   \vspace{-5mm}
% \end{table}

\begin{table}[t]
\small
\centering
\caption{Experiment Environment Setup. $R$ and $W$ represent \textit{Regular} and \textit{Weak} network condition, respectively.}
%\vspace{-2mm}
\resizebox{.4\textwidth}{!}{
  \begin{tabular}{c c  c c c c c}
    \toprule
     \parbox[c]{2.7cm}{\hrule height 0pt width 0pt \centering \textbf{Exp}}&
     \parbox[c]{0.5cm}{\hrule height 0pt width 0pt \centering \textbf{S1}}&
     \parbox[c]{0.5cm}{\hrule height 0pt width 0pt \centering \textbf{S2}}&
     \parbox[c]{0.5cm}{\hrule height 0pt width 0pt \centering \textbf{S3}}&
     \parbox[c]{0.5cm}{\hrule height 0pt width 0pt \centering \textbf{S4}}&
     \parbox[c]{0.5cm}{\hrule height 0pt width 0pt \centering \textbf{S5}}&
     \parbox[c]{0.5cm}{\hrule height 0pt width 0pt \centering \textbf{E}}\\
    \midrule
    \parbox[c]{2.7cm}{\centering A}& R & R & R & R & R & R\\   
    \parbox[c]{2.7cm}{\centering B}& R & W & R & W & R & W\\   
    \parbox[c]{2.7cm}{\centering C}& W & W & W & R & R & R\\   
    \parbox[c]{2.7cm}{\centering D}& W & W & W & W & W & W\\   
    \bottomrule
  \end{tabular}
  }
  \label{table:EE}
  \vspace{-1mm}
\end{table}
\subsection{Results}

%In this section, we demonstrate the efficiency 
We demonstrate the efficacy of our hybrid learning strategy for RL-based inference orchestration compared to two state-of-the-art RL-based inference orchestration~\cite{liu2020adadeep,kim2020autoscale}: 
AdaDeep~\cite{liu2020adadeep} employs the DQL algorithm to optimally orchestrate DL model selection based on available resources; 
AutoScale~\cite{kim2020autoscale} applies QL algorithm to optimally orchestrate DL inference in end-edge architecture (See Table~\ref{table:related}).
We evaluate the performance of the agent on a multi-user end-edge-cloud framework (see Section \ref{OnlineLearningFramework}). 
We also report the overhead incurred by the agent in our learning process, compared with AdaDeep~\cite{liu2020adadeep} and AutoScale~\cite{kim2020autoscale}.
\subsubsection{Agent Performance}
We demonstrate our proposed hybrid learning agent's performance in finding optimal orchestration decisions. 
At design time, we determine the true optimal configuration for orchestrating a DL task under any given condition of workloads, network, and number of active users using a brute force search.
This is used for 
%the evaluation purpose of 
comparing the orchestration decisions made by our proposed approach and DQL against the true optimal configuration.
Both Deep-Q Learning (DQL) and Hybrid Learning (HL) algorithms have yielded a 100\% prediction accuracy in comparison with the true optimal configuration. 
Thus, RL-based orchestration decisions always converge with the optimal solution. 
We investigate the agent's ability to find the optimal orchestration decisions under different scenarios of varying network conditions. 
Table \ref{table:EE} summarizes the experimental scenarios A-D with different combinations of regular (R) and weak (W) networks. 
For example, in experimental scenario A, all the nodes are connected with a regular network, whereas in scenario B, nodes $S1$, $S3$, and $S5$ have regular connections and the rest have weak connections. 
Orchestration decisions made by our proposed hybrid agent over four different experimental scenarios (A-D) are shown in Table \ref{tab:networkvariation}. 
For each end-user sensor device node (S1-S5) within each experimental scenario (A-D), we present the orchestration decisions viz., choice of execution node (among local device L, edge node E, cloud node C) and inference model (d0-d7, in decreasing order of accuracy) made by our hybrid agent. 
Table \ref{tab:networkvariation} also shows the average response time (ART, in ms) and average accuracy with the selected model (AA, in \%), along with the constraint (Cnst) on minimum accuracy requirement. 
Note that within each experimental scenario, the average response is lower as the accuracy threshold is relaxed. 
For instance, in experimental scenario A for device S1, models $d0$, $d4$,\;$d2$,\;$d7$ and $d7$ are selected respectively for accuracy thresholds ranging from \textit{Max} through \textit{Min}. 
Our proposed orchestrator explores the Pareto-optimal space of model selection and offloading choice of nodes to minimize latency within accuracy constraints. 
For instance in experimental scenario A, maintaining an accuracy level of 89\% results in an average response time of 269.8ms, by:
i) setting the models to $d4$,\;$d4$,\;$d4$,\;$d0$, and $d4$ on devices S1-S5, and 
ii) device configurations to L (local device), L, L, E (edge) and L for S1-S5. 
However, the average response time can be improved by sacrificing the accuracy within a predetermined tolerable level. 
For instance, by lowering the accuracy threshold by 4\% (from 89\% to 85\%), the average response time can be reduced by 46\% (from 269ms to 143ms) by:
i) setting the models to $d2$,\;$d6$,\;$d5$,\;$d6$, and $d5$ on devices S1-S5, and 
ii) device configurations to L (local device), L, L, L and L for S1-S5. 
With varying network conditions, our solution explores the offloading and model selection Pareto-optimal space to predict the optimal orchestration decisions.

\begin{table}[]
\caption{Results of the framework for different accuracy constraints for different experiments (five users). Cnst, ART, and AA represent constraint, average response time, and average accuracy, respectively. For example, in Exp-$D$ with $89\%$ average accuracy constraint, our framework orchestrates $S1$,\;$S2$,\;$S3$, and $S4$ to execute DL inference using model $d4$ locally and offload execution using model $d0$ at the cloud.}
\label{tab:networkvariation}
\centering
\resizebox{0.5\textwidth}{!}{
\begin{tabular}{@{}ccccccccc@{}}
\toprule
  &  & \multicolumn{5}{c}{\textbf{End-node Devices}} & \textbf{} &  \\ \midrule 
 \textbf{Exp} & \textbf{Cnst} & \textbf{S1} & \textbf{S2} & \textbf{S3} & \textbf{S4} & \textbf{S5} & \parbox[c]{0.5cm}{\hrule height 0pt width 0pt \centering\textbf{ART (ms)}} & \parbox[c]{0.5cm}{\hrule height 0pt width 0pt \centering\textbf{AA (\%)}} \\ \midrule
 \multirow{4}{*}{\textbf{A}} & Min & $d7,L$ & $d7,L$ & $d7,L$ & $d7,L$ & $d7,L$ & 72.08 & 72.80 \\
                                 & 80\%   &   $d7,L$      &   $d6,L$      &    $d6,L$     &    $d6,L$    &   $d6,L$     &   103.88       & 81.11 \\
                                 & 85\%   &   $d2,L$      &   $d6,L$      &    $d5,L$     &   $d6,L$     &   $d5,L$     &   143.81        & 85.06 \\
                                 & 89\%   &   $d4,L$      &   $d4,L$      &    $d4,L$     &   $d0,E$     &   $d4,L$     &   269.80        & 89.10 \\
                                 & Max   &   $d0,E$      &   $d0,L$      &   $d0,L$      &  $d0,C$      &   $d0,L$     &  418.91         & 89.90 \\
 \midrule 
 \multirow{4}{*}{\textbf{B}} & Min &   $d7,L$      &   $d7,L$      &   $d7,L$      &  $d7,L$      &  $d7,L$      &   106.76        & 72.80 \\
                                 & 80\%   &   $d6,L$      &   $d3,L$      &   $d6,L$      &   $d6,L$     &   $d6,L$     &   139.92        & 83.23 \\
                                 & 85\%   &   $d5,L$      &    $d5,L$     &   $d6,L$      &  $d6,L$      &  $d2,L$      &   176.21        &  85.05\\
                                 & 89\%   &   $d4,L$      &   $d4,L$      &   $d0,E$      &   $d4,L$     &  $d4,L$      &   303.50        & 89.10 \\
                                 & Max   &   $d0,C$     &  $d0,E$       &   $d0,L$      &  $d0,L$      &  $d0,L$       &   472.88       & 89.90 \\
 \midrule
 \multirow{4}{*}{\textbf{C}} & Min &   $d7,L$      &   $d7,L$      &  $d7,L$       &   $d7,L$     &  $d7,L$      &   119.28        & 72.80 \\
                                 & 80\%   &  $d6,L$       &   $d6,L$      &  $d7,L$       &  $d6,L$      &   $d6,L$     &   149.52        & 81.11 \\
                                 & 85\%   &   $d5,L$      &  $d6,L$       &   $d5,L$      &  $d6,L$      &  $d5,L$      &   190.76        & 85.47 \\
                                 & 89\%   &   $d4,L$      &  $d4,L$       &  $d4,L$       &  $d4,L$      &   $d0,C$     &   318.45        & 89.10 \\ 
                                 & Max   &   $d0,L$       &    $d0,L$      &     $d0,L$     &   $d0,C$      &   $d0,E$   &    464.59       & 89.90 \\ 

 \midrule
 \multirow{4}{*}{\textbf{D}} & Min &  $d7,L$       &  $d6,L$       &   $d7,L$      &   $d7,L$     &  $d7,L$      &   158.53        & 72.80 \\
                                 & 80\%   &  $d6,L$       &   $d6,L$      &   $d6,L$      &  $d7,L$      &  $d6,L$      &        182.53   & 81.12 \\
                                 & 85\%   &    $d2,L$     &   $d6,L$      &   $d6,L$      &  $d5,L$      &  $d5,L$      &   225.32        & 85.06 \\
                                 & 89\%   &  $d4,L$       &   $d4,L$      &   $d4,L$      &   $d4,L$     &  $d0,C$      &    356.75       & 89.10 \\ 
                                 & Max   &   $d0,L$      &   $d0,C$      &   $d0,E$      &   $d0,L$     & $d0,L$       &   506.62        & 89.90 \\ 

 \bottomrule
\end{tabular}}
\end{table}

\begin{figure}[t!]
\centering
\includegraphics[width=0.5\textwidth]{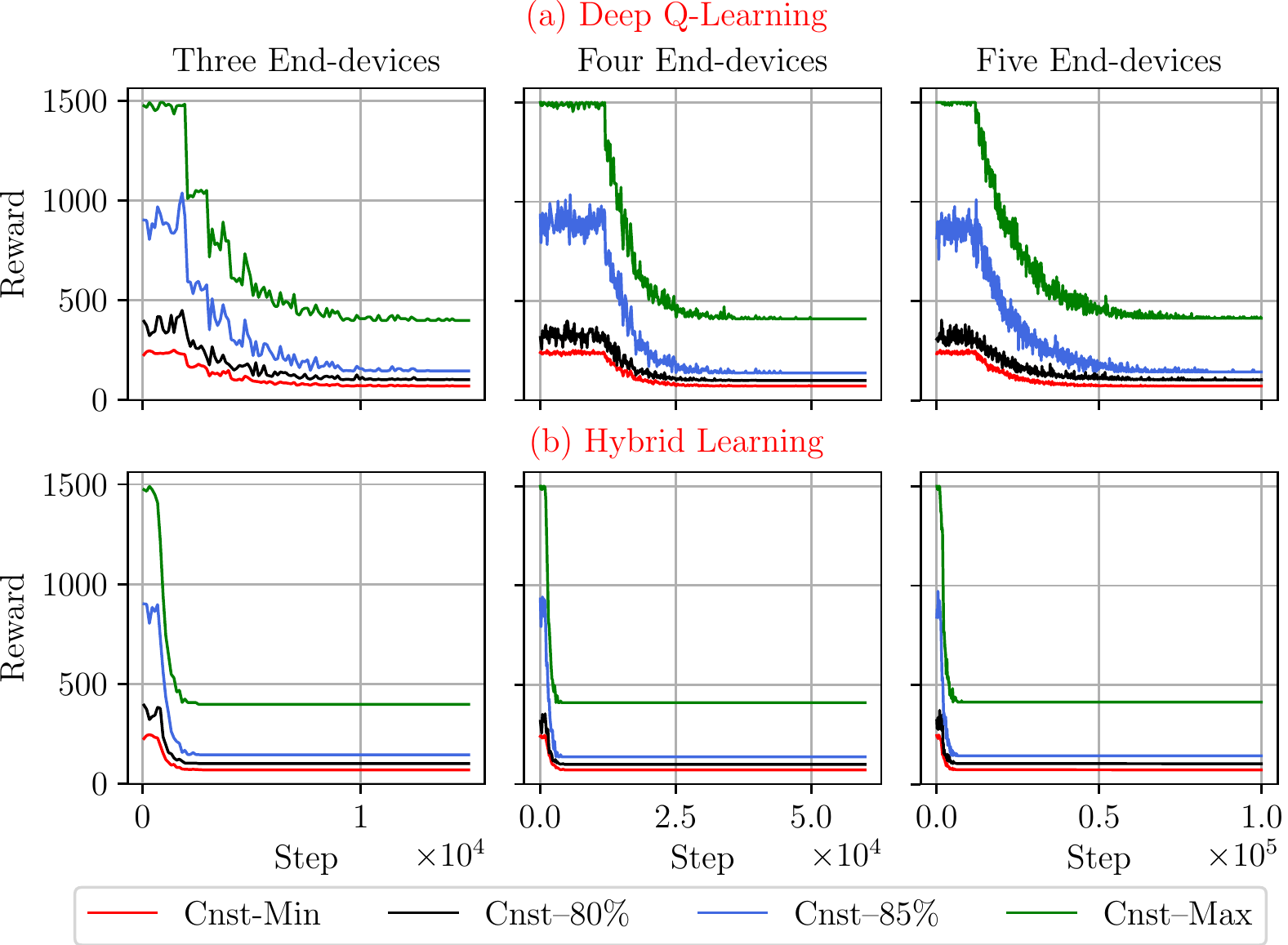}
\caption{Convergence time for up to five users within different constraints. Cnst represents constraint for each experiment. Deep Q-Learning referes to AdaDeep~\cite{liu2020adadeep} work. The comparison with AutoScale~\cite{kim2020autoscale} is mentioned in Table~\ref{tab:convergencetime}.}
\label{fig:hybridlearning}
\vspace{-6mm}
\end{figure}
\subsubsection{Training Overhead}
We evaluate the agent overhead during the training phase to demonstrate the efficiency of the hybrid learning algorithm in comparison with the state-of-the-art AutoScale~\cite{kim2020autoscale} and AdaDeep~\cite{liu2020adadeep}. 
To identify an optimal policy, we assess the number of steps required to interact with the system environment under each approach.
Figure \ref{fig:hybridlearning} shows the training phases for different number of users under different accuracy constraints. Each subplot shows the training phase for the system with a different number of users using the DQL and HL algorithms. The agent is trained under different accuracy constraints, which results in converging to different optimal policies for the corresponding constraint (i.e., different converged reward values). 
Our evaluation shows that the HL algorithm accelerates the training steps up to $11.6\times$ and $166.6\times$ in comparison with AdaDeep~\cite{liu2020adadeep} and AutoScale~\cite{kim2020autoscale} respectively. 
The convergence steps for different number of users are summarized in Table \ref{tab:convergencetime}. 
Our result shows that the number of agent's interactions with the system environment increases as we increase the dimension of the problem space (increasing number of users). 
However, our agent with the HL algorithm outperforms the state-of-the-art~\cite{kim2020autoscale,liu2020adadeep} in the number of interactions with the system environment for the different number of users and under different accuracy constraints. 
The training time consists of experience time (i.e., time spent in interacting with the system environment to collect data) and computation time for learning the system and policy models. 
Table \ref{tab:trainingtime} shows the overall training time for different number of users. 
Our evaluation shows that the HL algorithm converges up to $7.5\times$ faster in comparison with AdaDeep, and $109.4\times$ faster in comparison with A.. Further, we also present the overhead of the training agent in finding optimal orchestration decisions, with \textit{experience time} and \textit{computation time} metrics. 
\textit{Experience time} is the total time to execute all taken steps (cost of interaction with the system environment) to identify an optimal policy, while \textit{computation time} is the time to train the agent. The HL algorithm results in $4.4\times$ and $9.4\times$ speedup in comparison with AdaDeep for \textit{Computation Time} and \textit{Experience Time}, respectively. 
% In addition, Table \ref{tab:trainingtime} shows the computation time per step during the training phase for both DQL and HL algorithms. 
The computation time per step to train \textit{System Model} and \textit{Policy Model} with the HL algorithm is higher than the DQL and QL algorithms. However, our HL algorithm converges significantly faster than the algorithms (See Table \ref{tab:convergencetime} and Table \ref{tab:trainingtime}). 
Therefore, any additional computation cost per step is more than compensated by a significant reduction in the number of interactions to identify an optimal policy.

% \begin{table}[t!]
% \caption{Training convergence time for three, four, and five users with the DQL~\cite{liu2018demand} and HL algorithms.}
% \label{tab:convergencetime}
% \centering
% \resizebox{.48\textwidth}{!}{
% \begin{tabular}{@{}ccccc@{}}
% \toprule
% \parbox[c]{1.5cm}{\hrule height 0pt width 0pt \centering \textbf{\# of Users}} & \textbf{Constraint} & \textbf{AutoScale} &\textbf{HL} & \textbf{AdaDeep}  \\ 
% \midrule
% \multirow{4}{*}{\textbf{3}} &Min                 &&   $0.2\times10^4$   &   $0.1\times10^5$    \\
% &80$\%$                &&     $0.2\times10^4$        &    $0.1\times10^5$     \\
% &85$\%$                &&      $0.2\times10^4$       &    $0.1\times10^5$      \\
% &Max                 &&    $0.2\times10^4$         &     $0.1\times10^5$         \\

% \midrule
% \multirow{4}{*}{\textbf{4}} &Min                 &&      $0.3\times10^4$       &   $0.3\times10^5$  \\
% &80$\%$                &&       $0.4\times10^4$                &              $0.4\times10^5$                \\  
% &85$\%$                &&       $0.3\times10^4$              &                  $0.4\times10^5$             \\
% &Max                 &&   $0.3\times10^4$    &    $0.3\times10^5$              \\

% \midrule
% \multirow{4}{*}{\textbf{5}} &Min                 &&    $0.6\times10^4$        &   $0.6\times10^5$      \\
% &80$\%$                &&    $0.6\times10^4$  & $0.6\times10^5$  \\
% &85$\%$               &&    $0.6\times10^4$  &   $0.7\times10^5$          \\
% &Max                 &&    $0.6\times10^4$     &    $0.7\times10^5$      \\ \bottomrule
% \end{tabular}
% }
% \vspace{-2mm}
% \end{table}

\begin{table}[t!]
\caption{Training overhead for Hybrid Learning algorithm compared with AdaDeep~\cite{liu2020adadeep} and AutoScale~\cite{kim2020autoscale}. Training overhead is presented as number of steps to achieve the optimal policy.}
\label{tab:convergencetime}
\centering
\resizebox{.48\textwidth}{!}{
\begin{tabular}{@{}ccccc@{}}
\toprule
\parbox[c]{1.5cm}{\hrule height 0pt width 0pt \centering \textbf{\# of Users}} & \textbf{Constraint} & \textbf{AutoScale} &\textbf{AdaDeep} & \textbf{Our}  \\ 
\midrule
\multirow{4}{*}{\textbf{3}} &Min                 &$0.7\times10^4$&   $0.1\times10^5$   &   $0.2\times10^4$    \\
&80$\%$                &$0.5\times10^4$&     $0.1\times10^5$        &    $0.2\times10^4$     \\
&85$\%$                &$0.3\times10^4$&      $0.1\times10^5$       &    $0.2\times10^4$      \\
&Max                 &$0.7\times10^4$&    $0.1\times10^5$         &     $0.2\times10^4$         \\

\midrule
\multirow{4}{*}{\textbf{4}} &Min                 &$0.9\times10^5$&      $0.3\times10^5$       &   $0.3\times10^4$  \\
&80$\%$                &$0.8\times10^5$&       $0.4\times10^5$                &              $0.4\times10^4$                \\  
&85$\%$                &$0.4\times10^5$&       $0.4\times10^5$              &                  $0.3\times10^4$             \\
&Max                 &$0.9\times10^5$&   $0.3\times10^5$    &    $0.3\times10^4$              \\

\midrule
\multirow{4}{*}{\textbf{5}} &Min                 &$0.1\times10^7$&    $0.6\times10^5$        &   $0.6\times10^4$      \\
&80$\%$                &$0.1\times10^7$&    $0.6\times10^5$  & $0.6\times10^4$  \\
&85$\%$               &$0.6\times10^6$&    $0.7\times10^5$  &   $0.6\times10^4$          \\
&Max                 &$0.1\times10^7$&    $0.7\times10^5$     &    $0.6\times10^4$      \\ \bottomrule
\end{tabular}
}
\vspace{-3mm}
\end{table}

\begin{table}[]
\caption{Training time (presented in minutes) for different number of users compared with AutoScale~\cite{kim2020autoscale} and AdaDeep~\cite{liu2020adadeep}. Comp and Exp represent \textit{Computational Time} and \textit{Experience Time}. *AutoScale employs the QL algorithm which has a low computational overhead.}
\label{tab:trainingtime}
\resizebox{.48\textwidth}{!}{
\begin{tabular}{@{}cllll@{}}
\toprule
\multicolumn{1}{l}{\textbf{\# of Users}} & \textbf{Time (min)} & \textbf{AutoScale*} & \textbf{AdaDeep} & \textbf{Ours} \\ \midrule
\multirow{3}{*}{\textbf{3}} & Comp   & - & 1 & 1 \\
                            & Exp  & $1.5\times10^2$ & $6.8\times10^1$ & $2.6\times10^1$ \\
                            & Total & $1.5\times10^2$ & $6.9\times10^1$ & $2.7\times10^1$ \\ \midrule
\multirow{3}{*}{\textbf{4}} & Comp   & $0.3$ & $1.0\times10^1$ & $3.6$  \\
                            & Exp  & $3.7\times10^2$ & $1.1\times10^2$ &  $1.3\times10^1$\\
                            & Total & $3.7\times10^2$ & $1.2\times10^2$ & $1.6\times10^1$ \\ \midrule
\multirow{3}{*}{\textbf{5}} & Comp   & $1.0\times10^1$ & $1.5\times10^2$ & $3.4\times10^1$ \\
                            & Exp  & $5.8\times10^3$ & $1.8\times10^2$ & $1.9\times10^1$ \\
                            & Total & $5.8\times10^3$ & $3.3\times10^2$ & $5.3\times10^1$ \\ \bottomrule
\end{tabular}
}
\end{table}
\section{Conclusion}
%In this work, 
We presented a hybrid learning based framework for orchestrating deep learning tasks in end-edge-cloud architectures. 
Our proposed hybrid learning strategy requires fewer interactions with the real-time execution runs, converging to an optimal solution significantly faster than state-of-the-art model-free RL approaches. 
We deployed our proposed framework on enterprise AWS end-edge-cloud system for evaluating  MobileNet kernels. 
Our hybrid learning approach accelerates the training process by up to $166\times$ in comparison with the state-of-the-art RL-based DL inference orchestration, while making optimal orchestration decisions after significantly early convergence. 
Our future work will explore cross-layer opportunities and more hardware-friendly RL algorithms.
% \red{add a line on ongoing/future work...  to address any perceived shortcomings that reviewers might bring up.}
\bibliographystyle{ieeetr}
\bibliography{related}

\begin{thebibliography}{10}

\bibitem{schmidhuber2015deep}
J.~Schmidhuber, ``Deep learning in neural networks: An overview,'' {\em Neural
  networks}, 2015.

\bibitem{wang2020convergence}
X.~Wang {\em et~al.}, ``Convergence of edge computing and deep learning: A
  comprehensive survey,'' {\em IEEE Communications Surveys \& Tutorials}, 2020.

\bibitem{khelifi2018bringing}
H.~Khelifi {\em et~al.}, ``Bringing deep learning at the edge of
  information-centric internet of things,'' {\em IEEE Communications Letters},
  2018.

\bibitem{yousefpour2018all}
A.~Yousefpour {\em et~al.}, ``All one needs to know about fog computing and
  related edge computing paradigms: A complete survey,'' {\em Journal of
  Systems Architecture}.

\bibitem{shahhosseini2021exploring}
S.~Shahhosseini {\em et~al.}, ``Exploring computation offloading in iot
  systems,'' {\em Information Systems}.

\bibitem{shahhosseini2019dynamic}
S.~Shahhosseini {\em et~al.}, ``Dynamic computation migration at the edge: Is
  there an optimal choice?,'' in {\em GLSVLSI'19}.

\bibitem{kim2020autoscale}
Y.~G. Kim {\em et~al.}, ``Autoscale: Energy efficiency optimization for
  stochastic edge inference using reinforcement learning,'' IEEE, 2020.

\bibitem{sutton2018reinforcement}
R.~S. Sutton {\em et~al.}, {\em Reinforcement learning: An introduction}.
\newblock MIT press, 2018.

\bibitem{lu2020optimization}
H.~Lu {\em et~al.}, ``Optimization of lightweight task offloading strategy for
  mobile edge computing based on deep reinforcement learning,'' {\em Future
  Generation Computer Systems}, 2020.

\bibitem{liu2020adadeep}
S.~Liu {\em et~al.}, ``Adadeep: A usage-driven, automated deep model
  compression framework for enabling ubiquitous intelligent mobiles,'' 2020.

\bibitem{zhan2020deep}
W.~Zhan {\em et~al.}, ``Deep-reinforcement-learning-based offloading scheduling
  for vehicular edge computing,'' {\em Internet of Things Journal}, 2020.

\bibitem{lin2021computation}
B.~Lin {\em et~al.}, ``Computation offloading strategy based on deep
  reinforcement learning for connected and autonomous vehicle in vehicular edge
  computing,'' {\em Journal of Cloud Computing}, 2021.

\bibitem{wang2019computation}
J.~Wang {\em et~al.}, ``Computation offloading in multi-access edge computing
  using a deep sequential model based on reinforcement learning,'' {\em
  Communications Magazine}, 2019.

\bibitem{kim2021autofl}
Y.~G. Kim and C.-J. Wu, ``Autofl: Enabling heterogeneity-aware energy efficient
  federated learning,'' in {\em MICRO-54: 54th Annual IEEE/ACM International
  Symposium on Microarchitecture}, pp.~183--198, 2021.

\bibitem{peng2018deep}
B.~Peng {\em et~al.}, ``Deep dyna-q: Integrating planning for task-completion
  dialogue policy learning,'' {\em arXiv preprint arXiv:1801.06176}, 2018.

\bibitem{sutton1990integrated}
R.~S. Sutton, ``Integrated architectures for learning, planning, and reacting
  based on approximating dynamic programming,'' in {\em Machine learning
  proceedings 1990}, 1990.

\bibitem{zhang2018improved}
Z.~Zhang, ``Improved adam optimizer for deep neural networks,'' in {\em IWQoS},
  2018.

\bibitem{howard2017mobilenets}
A.~G. Howard {\em et~al.}, ``Mobilenets: Efficient convolutional neural
  networks for mobile vision applications,'' 2017.

\end{thebibliography}

\end{document}